\documentclass[aoas,nameyear,dvips]{arximspdf}
\usepackage{dcolumn,multirow}
\usepackage{graphicx}

\doi{10.1214/10-AOAS367}
\volume{4}
\issue{4}
\pubyear{2010}
\firstpage{2049}
\lastpage{2072}

\makeatletter
\newcolumntype{d}[1]{D{.}{.}{#1}}
\newtheorem{proposition}{Proposition}
\newtheorem{lemma}{Lemma}

\newcommand{\Y}{\mathbf{Y}}
\newcommand{\X}{\mathbf{X}}
\newcommand{\M}{\mathbf{M}}
\newcommand{\I}{\mathbf{I}}
\newcommand{\w}{\mathbf{w}}
\newcommand{\Q}{\mathcal{Q}}
\newcommand{\B}{\mathbf{B}}
\renewcommand{\S}{\mathbf{S}}
\renewcommand{\d}{\mathbf{d}}
\newcommand{\N}{\mathcal{N}_\I}

\makeatother

\begin{document}
\begin{frontmatter}

\title{Node harvest}

\runtitle{Node harvest}

\begin{aug}
\author[A]{\fnms{Nicolai} \snm{Meinshausen}\ead[label=e1]{meinshausen@stats.ox.ac.uk}\corref{}}

\runauthor{N. Meinshausen}
\affiliation{University of Oxford}

\address[A]{Department of Statistics\\
University of Oxford\\
1 South Parks Road\\
Oxford OX1 3TG\\
UK\\
\printead{e1}} 
\end{aug}

\received{\smonth{10} \syear{2009}}
\revised{\smonth{4} \syear{2010}}

%
\begin{abstract}
When choosing a suitable technique for regression and classification
with multivariate predictor variables, one is often faced with a
tradeoff between interpretability and high predictive accuracy. To give
a classical example, classification and regression trees are easy to
understand and interpret. Tree ensembles like Random Forests provide
usually more accurate predictions. Yet tree ensembles are also more
difficult to analyze than single trees and are often criticized,
perhaps unfairly, as `black box' predictors.

\textit{Node harvest} is trying to reconcile the two aims of
interpretability and predictive accuracy by combining positive aspects
of trees and tree ensembles. Results are very sparse and interpretable
and predictive accuracy is extremely competitive, especially for low
signal-to-noise data. The procedure is simple: an initial set of a few
thousand nodes is generated randomly. If a new observation falls into
just a single node, its prediction is the mean response of all training
observation within this node, identical to a tree-like prediction. A~new observation falls typically into several nodes and its prediction
is then the weighted average of the mean responses across all these
nodes. The only role of \textit{node harvest} is to `pick' the right
nodes from the initial large ensemble of nodes by choosing node
weights, which amounts in the proposed algorithm to a quadratic
programming problem with linear inequality constraints. The solution is
sparse in the sense that only very few nodes are selected with a
nonzero weight. This sparsity is not explicitly enforced. Maybe
surprisingly, it is not necessary to select a tuning parameter for
optimal predictive accuracy. \textit{Node harvest} can handle mixed
data and missing values and is shown to be simple to interpret and
competitive in predictive accuracy on a variety of data sets.
\end{abstract}



\end{frontmatter}

\section{Introduction}
Let $\Y=(\Y_1,\ldots,\Y_n)$ be a vector of $n$ observations of a
univariate real-valued response and $\X$ be the $n\times p$-dimensional
matrix, where the row-vector $\X_{i\cdot}\in\mathcal{X}$ is the
$p$-dimensional covariate for the $i$th observation for $i=1,\ldots,n$.
When trying to predict a new response, given covariates, regression
trees [\citet{CART}] are attractive since they are very simple to build
and understand. They are one example of a wider range of recursive
partitioning methods. For the sake of notational simplicity, let the
notion of a node in a tree and the corresponding subspace of
$\mathcal{X}$ be identical. Let $\Q$ be a collection of $q$ nodes,
where a node $Q_g\in\Q$, $g=1,\ldots,q$, is defined by a rectangular
subspace of $\mathcal{X}$,
\[
Q_g=\bigl\{\mathbf{x}\in\mathcal{X}\dvtx\mathbf{x}_k\in I^{(g)}_k\mbox{ for }k=1,\ldots,p\bigr\},
\]
and each interval $I^{(g)}_k$ is a subset of the support of the $k$th
covariate.

The leaf nodes of a tree form a partition of $\mathcal{X}$ in that
their union is identical to $\mathcal{X}$ and all pairwise
intersections are empty. If each leaf node is an element of
$\mathcal{Q}$, the partition corresponding to a tree can be expressed
by a weight vector $\w\in\{0,1\}^q$, where $\w_g=0$ means that node $g$
\textit{is not} used in the partition, while $\w_g=1$ means that node
$g$ \textit{is} used in the partition. The tree-style prediction
$\hat{Y}(\mathbf{x})$ at a point $\mathbf{x}\in\mathcal{X}$ is
then the observed mean
over all training observations in the same node,
%
\begin{equation}\label{eq:weighted}
\hat{Y}(\mathbf{x})=\sum_{g=1}^q \mu_g 1\{\mathbf{x}\in Q_g\}\w_g,
\end{equation}
where $\mu_g$ is the mean over all observations falling into node
$Q_g$,
\[
\mu_g=\frac{\sum_{i=1}^n 1\{\X_{i\cdot\in Q_g}\}\Y_i}{\sum_{i=1}^n 1\{\X_{i\cdot\in Q_g}\}} .
\]
The predictions on the $n$ observed samples can be conveniently written
as $\M\w$, where $\M$ is the $n \times q$-dimensional matrix, with row
entries for $i=1,\ldots,n$ given by
%
\begin{equation}\label{eq:M}
\M_{ig} =\cases{
\mu_{g},&\quad if $\X_{i\cdot}\in Q_g$\cr
0,&\quad if $\X_{i\cdot}\notin Q_g$}
\qquad
\mbox{for }g=1,\ldots,q=|\mathcal{Q}|.
\end{equation}
The empirical squared error
loss on the training samples is then
%
\begin{equation}\label{emploss}
\|\Y- \M\w\|^2
\end{equation}
and trees try to pick a partitioning by a tree (and a weight vector
$\w$ equivalently) that minimizes this empirical loss (\ref{emploss}),
under certain complexity constraints on the tree. These complexity
constraints can, for example, entail a penalty on tree size or a lower
bound on the number of observations in each node [\citet{CART}]; for an
alternative approach see \citet{blanchard2007optimal}. The optimal
values of the complexity constraints are typically determined by
cross-validation.

Compared to single regression trees, predictive accuracy is often
improved by tree ensembles. Boosting
[\citet{freund1996enb}; \citet{friedman00additive}], bagging
[\citet{breiman1996bagging}] and Random Forests [\citet{breiman01random}]
are popular techniques to create these ensembles. Predictions are
weighted averages over the output of all trees in the ensemble. They
thus effectively allow an observation to be part of more than one node.
For Random Forests [\citet{breiman01random}], each of $m$ trees in the
ensemble receives equal weight $1/m$. If all leaf nodes of the Random
Forest are part of the set $\mathcal{Q}$ above, the empirical loss can
again be written as in (\ref{emploss}) with the only difference that
now $\w_g \in\{0,1/m, 2/m,\ldots,1\}$ instead of the binary weights
$\w_g=\{0,1\}$ for trees. If a node appears only once in the ensemble,
its weight is $1/m$. If it appears more than once, the associated
weight is the corresponding multiple of $1/m$, up to a maximum of 1 if
the node appears in every tree of the ensemble.

Here, we explore the possibility of allowing arbitrary weights $\w_g
\in[0,1]$. Rather than growing trees greedily, we start from a large
set $\mathcal{Q}$ of potential nodes that are either obtained by random
splits or picked from an initial tree ensemble, just as in `Rule
ensembles' [\citet{friedman2005plv}]. While `Rule ensembles' uses the
nodes as binary indicator variable in a linear model with an
$\ell_1$-penalty on coefficients, \textit{node harvest} retains
tree-like predictions of the form (\ref{eq:weighted}). The only task of
\textit{node harvest} is finding suitable weights on nodes. Minimizing
the empirical loss (\ref{emploss}) under suitable constraints on the
weights turns out to be a quadratic program with linear inequality
constraints, which can be solved efficiently.

The goal of the proposed \textit{node harvest} procedure is two-fold:
On the one hand, a very competitive predictive accuracy (with
practically no adjustment of tuning parameters). On the other hand,
simple, interpretable results and predictions.

Random Forests satisfy the first of these demands but not necessarily
the latter since hundreds of large trees with thousands of nodes are
involved in the final decision. Marginal importance measures can be
calculated as proposed in [\citet{breiman01random}], but they only
describe some limited characteristics of the fitted function and
certainly do not explain the whole fit. Trees, on the other hand,
satisfy the second constraint but fall short of optimal predictive
accuracy. Moreover, if tree size is chosen by cross-validation, the
interaction order (tree depth) can be very high, lowering
interpretability. \textit{Node harvest} has the advantage of delivering
very accurate results while using in general only main effects and
two-factor interactions.

\textit{Node harvest} is introduced in Section~\ref{section:node}. An
extension to binary classification, dealing with missing values and
additional regularization of the estimator are covered in
Section~\ref{section:extensions}, while numerical results are shown in
Section~\ref{section:numerical}.

\section{Node harvest}\label{section:node}

\textit{Node harvest} (NH) is introduced, along with an efficient
algorithm to solve the involved quadratic programming problem. Some
basic properties of the estimator are established.

\subsection{Optimal partitioning}
The starting point of NH is loss function (\ref{emploss}). Suppose one
would like to obtain a partitioning of the space that minimizes the
empirical loss (\ref{emploss}). One could collect a very large number
of nodes into the set $\Q$ that satisfy desired complexity criteria.
Typical complexity criteria are a minimal node size or maximal
interaction depth (tree depth). An empirically optimal partitioning
would search for a weight vector such that the empirical loss is
minimal,
%
\begin{eqnarray}\label{optpart}
\hat{\w} = \operatorname{argmin}\limits_\w\| \Y -\M\w\|^2\nonumber
\\[-9.5pt]\\[-9.5pt]
\eqntext{\mbox{such that }\w\in\{0,1\}^q\mbox{ and }\{Q_g\dvtx \w_g =1\}\mbox{ is a partition of }\mathcal{X}.}
\end{eqnarray}
The selected set $\{Q_g: \w_g =1\}\subset\mathcal{Q}$ of nodes is
understood to form a partition iff the intersection between all
selected nodes is empty and their union is the entire space
$\mathcal{X}$. Even if given a collection $\mathcal{Q}$ of nodes, the
optimization problem above is very difficult to solve. The constraint
$\w\in\{0,1\}^q$ does not correspond to a convex feasible region.
Moreover, the constraint that the selected set of nodes form a
partition of the space is also awkward to handle computationally.

The latter problem can be circumvented by demanding instead that the
partition is a proper partitioning for the \textit{empirically observed
data only} in the sense that each data point is supposed to be part of
exactly one node. This loosening of the constraint will be very helpful
at a later stage. It might create the situation that a new observation
will not belong to any node, but this will turn out to be not a problem
in the NH approach since every observation will be a member of the root
node and the root node always receives a small positive weight, which
is discussed further below.

To form such an empirical partitioning, let $\I$ be the $n \times q$
matrix indicating whether or not an observation falls into a given
leaf. For all rows $i=1,\ldots,n$,
%
\begin{equation} \label{I}
\I_{ig}=\cases{
1,&\qquad if $\X_{i\cdot}\in Q_g$\cr
0,&\qquad if $\X_{i\cdot}\notin Q_g$
}
\qquad\mbox{ for }g=1,\ldots,q.
\end{equation}
The constraint that each data point be
part of one and exactly one node is equivalent to demanding that $\I\w
=1$, understood componentwise. Since $\w\in\{0,1\}^q$, this simple
linear equality constraint ensures that each observation is part of
exactly one selected node.

Given a collection $\Q$ of nodes, a weight vector $\w$ could thus be
found by the constrained optimization
%
\begin{equation}\label{eq:what01}
\hat{\w}=\operatorname{argmin}\limits_{\w}\|\Y-\M\w\|^2 \qquad\mbox{such that }\I\w=1\mbox{ and }\w\in\{0,1\}^q.
\end{equation}
For the $n$ observed data points, this
problem is equivalent to (\ref{optpart}), yet it is still NP-hard to
solve in general due to the nonconvex feasible region of the constraint
$\w\in\{0,1\}^q$. Tree ensembles relax this constraint and average
over several trees, implicitly allowing weights to take on values in
the interval $[0,1]$. It thus seems natural to relax the nonconvex
constraint $\w\in\{0,1\}^q$ and only ask for nonnegativity of the
weights.

\subsection{Node harvest}
The main idea of NH is that it becomes computationally feasible to
solve the optimal empirical partitioning problem (\ref{eq:what01}) if
the weights are only constrained to be nonnegative. The weights across
all nodes for a single observation still have to sum to 1 (as they do
for all weighted tree ensembles), but this constraint is equivalent to
$\I\w=1$, and we can relax (\ref{eq:what01}) to the convex
optimization problem
%
\begin{equation}\label{eq:what}
\hat{\w}=\operatorname{argmin}\limits_{\w}\|\Y-\M\w\|^2\qquad\mbox{such that }\I\w=1\mbox{ and }\w\ge 0.
\end{equation}
This estimator is called the \textit{node harvest}
(NH) estimator since a small subset of nodes is `picked' or selected
from a large initial ensemble of nodes. It will turn out that the vast
majority of nodes in this large ensemble will receive a zero weight,
without the sparsity being enforced explicitly other than through the
constraint $\I\w=1$. Nodes $g$ which receive a zero weight
($\hat{\w}_g=0$) can be ignored for further analysis.

The constraints in (\ref{eq:what}) are satisfied, for example, by
setting the weight of the root node, which is always included in $\Q$
and contains all observations, equal to 1 and all other weights to 0.
The set of solutions is thus always nonempty. The solution to
(\ref{eq:what}) is also either unique or the set of solutions is a
convex set. In the latter case, we define $\hat{\w}$ for definiteness
to be the solution that has minimal $\ell_2$-norm among all solutions
in this convex set, which amounts to adding a small ridge penalty $\nu
\|\w\|_2^2$ to the objective function in (\ref{eq:what}) and letting
$\nu\to0$. Other solutions are possible, but adding a very small ridge
penalty guarantees, moreover, positive definiteness of the quadratic
form and facilitates computation of (\ref{eq:what}) even if the
solution is unique.

The prediction for new data is then simply a weighted average over node
means. For the training data, this is still the vector $\M\w$. The
prediction $\hat{Y}(\mathbf{x})$ for a new data point $\mathbf{x}\in
\mathcal{X}$ is
the weighted average over all nodes that $\mathbf{x}$ falls into,
%
\begin{equation}\label{weightedmean}
\hat{Y}(\mathbf{x})=\frac{\sum_{g\in G_\mathbf{x}}\hat{\w}_g\mu_g}{\sum_{g\in G_\mathbf{x}}\hat{\w}_g},
\end{equation}
where $G_\mathbf{x}:=\{g\dvtx\mathbf{x}\in Q_g\}$ is the collection of
nodes that
observation $\mathbf{x}$ falls into.

The denominator in (\ref{weightedmean}) is constrained to be 1 for all
$n$ training samples since $\I\w=1$ is enforced. For new observations
outside the training set, the weights in the denominator do not
necessarily sum to 1. We always let the root node be a member of the
set $\Q$, where the root node is defined as containing the entire
predictor space~$\mathcal{X}$. We demand that the weight of the root
node is bounded from below not by 0 as for all other nodes, but by a
very small weight chosen here as $0{.}001$ and converging to 0 for
increasing sample sizes. The set $G_\mathbf{x}$ in (\ref
{weightedmean}) is
then always nonempty and the denominator in (\ref{weightedmean}) is
bounded from below by $0{.}001,$ although it will typically be in the
region of 1 for new observations. In the unlikely event that a new
observation is not part of any node except the root node, its
prediction will, according to (\ref{weightedmean}), be the node mean of
the root node. This is identical to the mean response over all
observations in the training data, a reasonable answer if a new
observation should fail to fall into any selected node.

\subsection{Tuning parameters}

The NH procedure requires only an initial set of nodes $\Q$. Once this
set is specified, there are no further tuning parameters. It will turn
out that results are very insensitive to the actual choice of the set
of nodes as long as $q=|\mathcal{Q}|$ is sufficiently large and some
complexity constraints, such as maximal interaction order and minimal
node size, are followed.

There are three essential characteristics of the set $\mathcal{Q}$:
the number of nodes, maximal interaction order and minimal node size.
We discuss these constraints in the following, but an advantageous
aspect of the proposed method is that the method is competitive in
terms of predictive accuracy for the default choices proposed below. In
fact, all numerical results are computed with the same defaults
parameters for maximal interaction order, which is set to 1, and
minimal node size, which is set to 5.

\textit{Number of nodes}. It will be shown empirically for many data
sets that the performance is continuously improving the more nodes
$q=|\mathcal{Q}|$ are added to the initial set of nodes. Solving (\ref
{eq:what}) gets clearly more costly as
$q$ increases. One should thus use as many nodes as can be afforded
computationally. Typically, $q$ ranges in the hundreds or thousands.
All examples are calculated with $q=1000$ nodes. It is maybe surprising
that there is practically no overfitting if $q$ is chosen very large. A
first attempt at explaining this phenomenon can be found in Proposition
\ref{propos:1}.

\textit{Maximal interaction order}. The maximal interaction order of
node $Q_g$ is the number of variables that are necessary to determine
whether an observation is part of a node or not. Main effects have thus
an interaction order 1.
To keep results as interpretable as
possible, a maximal interaction order of 2 (equivalent to a two-factor
interaction) is chosen for almost all examples.

\textit{Minimal node size}. The minimal node size $\min_g|\{i\dvtx\X_{i\cdot}\in Q_g\}|$ has an influence on the amount of
smoothing. Allowing nodes with just a single observation, the algorithm
could simply interpolate all observed data by assigning weights of 1 to
the $n$ nodes that contain each exactly one of the $n$ observations.
This is clearly undesirable and a minimal node size of 5 is imposed
throughout. Again, results could be improved for some data sets by
tuning this choice, yet the results show that a choice of 5 gives very
competitive results across a remarkably wide range of data sets.

\subsection{Node generation}
To generate the desired nodes, one can generate nodes at random,
without use of the response variable. Alternatively, one can use a
data-adaptive choice by using nodes from a fitted tree ensemble.
Results seem very insensitive to this choice, but the latter method
requires in general fewer nodes in the initial set $\mathcal{Q}$ for a
close to optimal predictive accuracy. We thus follow the latter
approach. The set $\mathcal{Q}$ is initially empty. A new tree is
grown as proposed in \citet{breiman01random} for each tree in a Random
Forest (RF) ensemble. To speed up computation and increase diversity
of the set, the trees are fitted on subsamples of the data of size
$n/10$ rather than bootstrap samples. All the nodes of the tree that
satisfy the maximal interaction order and minimal node size constraint
are added to the set~$\mathcal{Q}$, provided that they are not already
present in the set. While the size of $\mathcal{Q}$ is less than the
desired number $q$, the procedure is repeated. If two or more nodes in
$\mathcal{Q}$ contain exactly the same set of training observations,
only a randomly chosen one of them is kept.

\subsection{Algorithm and dimensionality reduction}
As stated above, the initial set of nodes $\mathcal{Q}$ is generated
with a Random Forests approach. After the desired number~$q$ of nodes
have been obtained, it only remains to solve (\ref{eq:what}). This is a
quadratic program (QP) with linear constraints and could be solved with
standard QP solvers. However, the specific structure of the problem can
be used to reduce dimensionality and make the computation more
efficient.

We suppose that the root node, containing all observations, is the
first among all $q=|\Q|$ nodes. Let $\w_{\mathrm{root}}$ be the vector
$\w_{\mathrm{root}}=(1,0,0,\ldots,0)$. Clearly, $\I\w_{\mathrm{root}} =1$ componentwise,
so the equality constraint in (\ref{eq:what}) is fulfilled for
$\w_{\mathrm{root}}$. This means that the difference $\hat{\w}-\w_{\mathrm{root}}$
between the actual solution and the `root' solution $\w_{\mathrm{root}}$ lies in
the nullspace $\N\subseteq\mathbb{R}^q$ of $\I$. Let $\tilde{q}$ be
the dimension of $\N$. Since $\I$ is of rank at most $\min\{q,n\}$, we
have $\tilde{q}\ge q- \min\{q,n\}$, and the nullspace $\N$ is
guaranteed to be nontrivial ($\tilde{q}>0$) for $q>n$, that is, if
there are more nodes than actual observations, which we can always
satisfy by generating sufficiently many nodes. If the nullspace is
nontrivial, then let $\B$ be the $q\times\tilde{q}$-dimensional
matrix, where the $k$th column, with $k=1,\ldots,\tilde{q}$, contains
the $k$th basis vector of an arbitrarily chosen orthonormal basis of
$\N$. The solution to (\ref{eq:what}) can then be written, using the
argument above, for some $\hat{\d}\in\mathbb{R}^{\tilde{q}}$ as $
\hat{\w} = \w_{\mathrm{root}} + \B\hat{\d}$, and, to get the same solution as
in (\ref{eq:what}), $\hat{\d}$ is the solution to
%
\begin{eqnarray}\label{eq:dhat}
\hat{\d} = \operatorname{argmin}\limits_{\d}-2\d^T (\M\B)^T(\Y-\overline{\Y})+\d^T (\M\B)^T(\M\B)\d\nonumber
\\[-9.5pt]\\[-9.5pt]
\eqntext{\mbox{ such that }\B\d\ge-\w_{\mathrm{root}},}
\end{eqnarray}
where it was used that $\M\w_{\mathrm{root}}=\overline{\Y}$ by
definition of $\w_{\mathrm{root}}$. If a small ridge penalty $\nu\|\w\|_2^2$ on
$\w$ is added to guarantee uniqueness of the solution, a term $\nu\|
(\w_{\mathrm{root}} + \B\d)\|_2^2$ is added to the objective function in
(\ref{eq:dhat}), where here always $\nu=0{.}001$ under a standardized
response with $\operatorname{Var}(\Y)=1$. To also ensure that the weight of
the root node is bounded from below by the small chosen value $0{.}001$
instead of 0, the constraint $\B\d\ge-\w_{\mathrm{root}}$ in (\ref{eq:dhat})
needs to be replaced by $\B\d\ge- 0.999 \w_{\mathrm{root}}$.

Thus, the original $q$-dimensional problem is reduced to a
$\tilde{q}\ge q-\min\{q,n\}$-dimensional one. A price to pay for
this is the computation of a basis for the nullspace $\N$ of $\I$,
which is achieved by a SVD of $\I$. Compared to the savings in the QP
solution, computation of the SVD is, however, very much worthwhile. The
remaining QP problem (\ref{eq:dhat}) is solved with the QP solver of
\citet{goldfarb1983numerically}, as implemented in the package
\texttt{quadprog} of the \texttt{R}-programming language [\citet{R}]. It
is conceivable that an alternative interior-point algorithm and
especially explicit use of the sparse structure of the matrixes $\M$
and $\I$ would generate additional computational savings, but, even so,
it took less than 10 seconds to solve (\ref{eq:dhat}) on data sets with
less than $10^3$ observations, using a 2.93 GHz processor and 8 GB of
RAM.

\subsection{Smoothing}

NH can be seen as a smoothing operation in that $\hat{\Y} =\S\Y$
for a
data-adaptive choice of the smoothing matrix $\S$. The smoothing matrix
is doubly stochastic, symmetric and has nonnegative entries.

\begin{lemma}\label{lemma:1}
The fitted values $\hat{\Y}$ are obtained as a linear transformation
$\hat{\Y} =\S\Y$ of the original data, where $\S$ is a doubly
stochastic and symmetric matrix in that $\sum_{j} \S_{ij} =1$ for all
$i=1,\ldots,n$ and $\sum_i \S_{ij}=1$ for all $j=1,\ldots,n$. Moreover,
$\S_{ij} \ge0$ for all $i,j=1,\ldots,n$.
\end{lemma}

\begin{pf}
The fitted values are for the $n$ training observations given by
$\hat{\Y}= \M\hat{\w}$, with $\M$ defined in (\ref{eq:M}). Therefore,
$\hat{\Y}_i = \sum_{ g =1}^q 1\{ i\in Q_g\} \hat{\w}_g \mu_g$, where
$i\in Q_g$ is a shorthand notation for $\X_{i\cdot}\in Q_g$. Let
$n_g=|\{j\dvtx j\in Q_g\}|$ be the number of samples in node $g$. Then
$\mu_g = n_g^{-1} \sum_{j \in Q_g} \Y_j$ by definition of the node
means and, hence, putting together,
\[
\hat{\Y}_i=\sum_{g=1}^q 1\{i\in Q_g\}\hat{\w}_g n_g^{-1}\sum_{j=1}^n 1\{j\in Q_g\}\Y_j=\sum_{j=1}^n\sum_{g=1}^q\frac{\hat{\w}_g 1\{i,j\in Q_g\}}{n_g}\Y_j.
\]
Defining matrix $\S\in\mathbb{R}^{n\times n}$ by its\vspace*{-4pt}
entries $\S_{ij} = \sum_{g} \hat{\w}_g n_g^{-1} 1\{ i,j \in Q_g\}
$, it
follows that (a) $\hat{\Y} =\S\Y$, (b) $\S$ is symmetric and (c) that
all entries are nonnegative. It remains to show that $\sum_j
\S_{ij}=1$ for all $i=1,\ldots,n$. The column sums follow by symmetry.
Now, $\sum_j \S_{ij} = \sum_j \sum_{g} \hat{\w}_g n_g^{-1} 1\{
i,j \in
Q_g\} = \sum_{g} \hat{\w}_g 1\{i\in Q_g\} $. By definition of the
matrix $\I$, the right-hand side $\sum_{g} \hat{\w}_g 1\{i\in Q_g\} $
is identical to the $i$th coefficient in $\I\hat{\w}$. Since,
componentwise, $\I\hat{\w}=1$ by (\ref{eq:what}), it follows that
indeed $\sum_j \S_{ij}=1$ for all $i=1,\ldots,n$, which completes the
proof.
\end{pf}

From the lemma above, one can immediately derive that the mean
$\overline{\hat{\Y}}$ of the fitted values is identical to the mean
$\overline{\Y}$ of the observed values. And the lemma above also
ensures that, irrespective of the size $q$ of the initial ensemble, it
is impossible to fit the response exactly by interpolation if the
minimal node size is strictly larger than 1.

\begin{proposition}\label{propos:1}
The mean of the fitted and observed values agree, $\overline{\hat{\Y}}
= \overline{\Y}$. Moreover, if the minimal node size is larger than 1,
the weight of the root node is strictly positive and
$\operatorname{Var}(\Y)\neq0$, it holds for any strictly convex real-valued
function $f$ that
%
\begin{equation}\label{eq:shrunken}
\sum_{i=1}^n f(\hat{\Y}_i)<\sum_{i=1}^n f(\Y_i) .
\end{equation}
\end{proposition}

\begin{pf}
The first claim follows directly from Lemma \ref{lemma:1} since
$\hat{\Y} =\S\Y$ and, hence, $\sum_{i=1}^n \hat{\Y}_i = \sum_{i,j=1}^n
\S_{ij} \Y_j = \sum_{j=1}^n \Y_j $, where the last equality follows by
the fact that $\sum_i \S_{ij}=1$ for all $j=1,\ldots,n$ from Lemma
\ref{lemma:1}. Likewise, observe that $\S_{ij} <1$ for all
$i,j=1,\ldots,n$ if the minimal node size is larger than 1. This
follows from the definition of $\S$ by the entries $\S_{ij} = \sum_{g}
\hat{\w}_g n_g^{-1} 1\{ i,j \in Q_g\}$ since more than 1 entry in each
row-vector $\S_{i\cdot}$, $i=1,\ldots,n$, has to be nonzero. Since the
sum of the row is constrained to $\sum_j \S_{i j}=1$ and all entries in
$\S$ are nonnegative, all entries have got to be strictly less than~1.
Moreover, if the weight of the root node is positive, all entries
$\S_{i,j}$ are strictly positive. Hence, for a strictly convex function
$f$,
\[
\sum_{i=1}^n f(\hat{\Y}_i)=\sum_{i=1}^n f\Biggl(\sum_{j=1}^n\S_{ij}\Y_j\Biggr)<\sum_{i=1}^n \sum_{j=1}^n\S_{ij}f(\Y_j)=\sum_{j=1}^n f(\Y_j),
\]
having used $\operatorname{Var}(\Y)\neq0$ and the
strict positivity of all entries of $\S$ in the inequality and $\sum_j
\S_{ij}=1$ for all $i=1,\ldots,n$, from Lemma~\ref{lemma:1} in the
last equality.
\end{pf}

The second part of the result can be obtained if the condition that the
weight of the root nodes is positive is replaced with the following
weaker condition: there exists a pair of observations $\Y_i,\Y_j$ with
$\Y_i\neq\Y_j$ such that both $i$ and $j$ are members of a node $\Q_g$
and the weight $\hat{\w}_g$ is strictly positive.

The proposition implies that the observed data cannot be interpolated
exactly by NH even though the number $q$ of nodes might greatly exceed
sample size $n$.

\subsection{Related work}
There has been substantial interest in the Random Forests framework for
classification and regression [\citet{breiman01random}], which builds
partly upon the randomized tree idea in \citet{amit97shape}.
\citet{lin2006random} interpreted Random Forests as an adaptive nearest
neighbor scheme, with the distance metric given by the grown tree
ensemble. The same interpretation is maybe even more imminent for NH
since predictions are explicitly averages over node means. Both
bagging [\citet{breiman1996bagging}] and boosting
[\citet{freund1996enb}; \citet{friedman00additive}] are possible alternative and
powerful techniques for growing multiple trees. If using either of
these, predictions are formed by averaging in a possibly weighted form
across all grown trees. Results are often difficult to interpret,
though, as each of possibly hundreds of grown trees consists in turn of
multiple nodes and all variables in the data set are often involved
somewhere in the ensemble. The influence of individual variables can
only be measured indirectly for such tree ensembles; see
\citet{strobl2007bias} for a more involved discussion. Despite a similar
sounding name, `tree harvesting' [\citet{hastie2001supervised}], a
regression technique commonly used in computational biology, is not
closely related to NH. An interesting machine learning technique is
`stacking' [\citet{wolpert1992stacked}; \citet{breiman1996stacked}], which is
weighting various classifiers and weights are chosen by minimizing the
error on weighted leave-one-out predictions. In contrast to stacked
trees, however, NH is not weighting whole trees but is working at the
level of individual nodes by reweighting each node. In a similar
spirit, the `Rule Ensemble' algorithm by \citet{friedman2005plv}
simplifies interpretability of tree ensembles by selecting just a few
nodes across all trees. Each node is seen to form a binary indicator
variable and the prediction is a linear combination of all these
indicator variables. In fact, for a given collection $\mathcal{Q}$ of
nodes, the matrix whose columns correspond to the binary indicator
variables is exactly the matrix defined as $\I$ in (\ref{I}). The
linear combination $\beta$ of nodes is then sought in a Lasso-style way
by putting a constraint on the $\ell_1$-norm of the coefficient vector
[\citet{tibshirani96regression}; \citet{chen01atomic}],
%
\begin{equation}\label{rules}
\hat{\beta}^\lambda=\operatorname{argmin}\limits_\beta\|\Y-\I\beta\|^2\qquad\mbox{ such that }\|\beta\|_1\le\lambda.
\end{equation}
The original variables can be added to the matrix $\I$ of binary
indicator variables. Despite the superficial similarity of `Rule
Ensembles' with NH, there are fundamental differences to the NH
procedure (\ref{eq:what}). Choosing the right tuning parameter~$\lambda$ is essential in (\ref{rules}), but no such tuning is
necessary for NH. The inherent reason for this is that NH imposes much
stronger regularization by requiring in (\ref{weightedmean}) that
predictions are weighted node means. NH is only selecting the weights
$\w$ in (\ref{eq:what}), whereas the vector $\beta$ in (\ref{rules})
cannot be interpreted as the weight attached to a particular node or
rule. The sign and magnitude of the coefficient $\beta_g$ is thus not
directly related to the average response of observations in node $g$. A
possible advantage of NH is thus the interpretability of the
predictions as weighted node means. An example is shown in the breast
cancer example in Figure~\ref{fig:breast}. If a new patient falls into
only a single node, the NH prediction is simply the average response in
the group of patients, which is very easy to communicate and relate to
the actually observed data. If he or she falls into several groups, the
prediction is the weighted average across these groups. In terms of
predictive power, rule ensembles seem to be often better than NH and
also Random Forests in our experience if the signal-to-noise is high
[\citet{meinshausen2009forest}]. The strength of NH is its ability to
cope well with very low signal-to-noise ratio data and the two
approaches seem complementary in this regard. Both `Rule Ensembles' and
NH can make use of a dictionary of rules, which is currently built
either randomly or by harvesting nodes from existing tree ensembles
such as Random Forests. More general nodes, such as spheres under
various metrics that are centered at training observations, could
conceivably help improve both methods.

\section{Extensions}\label{section:extensions}
\textit{Node harvest} (NH) can be extended
and generalized in various ways, as briefly outlined below. NH is
shown to be directly applicable to binary classification. Missing
values can easily be dealt with, without using imputation techniques
or surrogate splits when predicting the response for new observations
with missing values. Finally, a regularization is proposed that can
reduce the number of selected nodes.

\subsection{Classification}

For binary classification with $Y\in\{0,1\}$, the nonconvex
misclassification loss is typically replaced with a convex majorant of
this loss function [\citet{bartlett03convexity}]. One of these possible
convex loss functions is the $L_2$-loss, as used for classification in
\citet{yu2003blr}.

Simply applying the previous QP problem (\ref{eq:what}) on binary data
leads to a prediction $\hat{Y}(\mathbf{x})$ at a new data point
$\mathbf{x}$ which is
identical to (\ref{weightedmean}). The node means $\mu_g$,
$g=1,\ldots,q$, are now equivalent to the fraction of samples in class
``1'' among all samples in node $Q_g$,
\[
\mu_g=\frac{|\{i\dvtx\X_{i\cdot}\in Q_g\mbox{ and }\Y_i=1\}|}{|\{i\dvtx\X_{i\cdot}\in Q_g\}|} .
\]
The NH predictions are naturally in the interval $[0,1]$. Use of the
$L_2$-loss as a convex surrogate for misclassification error is thus
not only appropriate for NH, it is even beneficial since it allows for
an interpretation of the predictions $\hat{Y}(\mathbf{x})$ as weighted
empirical node means.

\subsection{Missing values}
An interesting property of NH is its natural ability to cope with
missing values. Once a fit is obtained, predictions for new data can be
obtained without use of imputation techniques or surrogate splits. To
fit the \textit{node harvest} estimator with missing data, we replace
missing values in the matrix $\X$ by the imputation technique described
in \citet{breiman01random} and \citet{liawclassification} and proceed
just as previously.

Suppose then that the \textit{node harvest} estimator is available and
one would like to get a prediction for a new observation $\X_{i\cdot}$
that has missing values in some variables. We still calculate the
prediction as the weighted mean (\ref{weightedmean}) over all nodes of which the
new observation is a member. The question is whether observation $i$
is part of node $Q_g\in\mathcal{Q}$ if it has missing values in
variables that are necessary to evaluate group membership of node
$Q_g$. The simplest and, as it turns out, effective solution is to say
that $i$ is not a member of a node if it has missing values in
variables that are necessary to evaluate membership of this node. To
make this more precise, let $Q_g$ be a node
\[
Q_g=\bigl\{\mathbf{x}\in\mathcal{X}\dvtx\mathbf{x}_k\in I^{(g)}_k\mbox{ for all }k\in\{1,\ldots,p\}\bigr\},
\]
and let $\mathcal{K}_g\subseteq\{1,\ldots,p\}$ be the set of variables
that are necessary and sufficient to evaluate node membership
[sufficient in the sense that $I^{(g)}_k$ is identical to the entire
support of $\mathbf{x}_k$ for all $k\notin\mathcal{K}_g$ and
necessary in the
sense that $I^{(g)}_k$ is \textit{not} identical to the support of
$\mathbf{x}_k$ for all $k\in\mathcal{K}_g$]. If $\mathbf{x}$ has
missing values, we
define
\[
\mathbf{x}\in Q_g\mbox{ if and only if, for all }k\in\mathcal{K}_g,\mathbf{x}_k\mbox{ is not missing and }\mathbf{x}_k\in I^{(g)}_k.
\]

Since we usually only work with main effects and two-factor
interactions, all nodes require only one or two variables to evaluate
node membership. Even with missing values in $\X_{i,\cdot}$,
observation~$i$ can still be a member of many nodes in $\mathcal{Q}$,
namely, those that involve only variables where the $i$th observation
has nonmissing values. In the most extreme case, \textit{all}
variables are missing from a new observation. The observation will then
\textit{only} be a member of the root node and the prediction is the
node mean of the root node, which is the mean of the response variable
across all training observations, maybe not an unreasonable answer in
the absence of any information. In more realistic cases, the new
observation will have \textit{some} nonmissing variables and be a
member of more than the root node and the prediction will be more
refined. With trees, a similar idea would amount to dropping a new
observation down a tree and stopping at the first node where the
split-variable is missing. The prediction would then naturally be the
mean response of observations within this node. However, if the
variables on which the root node is split are missing, the predicted
response will be the mean across all observations. This situation
occurs for NH only typically if all variables are missing. The
use of surrogate variables [\citet{CART}] is thus paramount for trees,
while NH can take a more direct approach.

\subsection{Regularization}

There is so far no tuning parameter in the NH procedure apart from the
choice of the large initial set $\mathcal{Q}$ of nodes. And results
are rather insensitive to the choice of $\mathcal{Q}$ as long as it is
chosen large enough, as shown in the next section with numerical
results.

Even though often not necessary from the point of predictive accuracy,
the method can be regularized to further improve interpretability. Here
it is proposed to constrain the average number of samples in each node.
From the outset, the minimal node size of 5 ensures that the average
fraction of samples in each node is above $5/n$. Even so, one might not
like to select many nodes that contain only a handful of observations.
The fraction of samples in node $g$ is $n_g/n$ and the weighted mean
across all nodes is
%
\begin{equation} \label{weightedfraction1}
\frac{\sum_g\hat{\w}_g(n_g/n)}{\sum\hat{\w}_g} ,
\end{equation}
where $n_g=|\{j\dvtx j\in Q_g\}|$ is again the number of samples in node
$g$. Since $\I\hat{\w}=1$ by (\ref{eq:what}), we have, by summing over
the rows of this equality,
\[
n=\sum_{i=1}^n\sum_{g=1}^q\I_{ig}\hat{\w}_g=\sum_{g=1}^q\hat{\w}_g\sum_{i=1}^n\I_{ig}=\sum_{g=1}^q\hat{\w}_g n_g ,
\]
where the last equality stems from the definition of matrix $\I$ in
(\ref{I}). The nominator in (\ref{weightedfraction1}) is thus 1 and
the weighted average fraction of samples (\ref{weightedfraction1})
within nodes is, maybe surprisingly, equal to the inverse of the
$\ell_1$-norm of the weight vector $\hat{\w}$,
%
\begin{equation}\label{weightedfraction}
\frac{\sum_g\hat{\w}_g(n_g/n)}{\sum_g \hat{\w}_g}=\frac{1}{\sum_g\hat{\w}_g}=\|\hat{\w}\|^{-1}_1.
\end{equation}
Constraining the $\ell_1$-norm of $\hat{\w}$ to be less than a positive
value of $\lambda\in[1,\infty]$ constrains thus the average fraction of
samples (\ref{weightedfraction}) to be at least $1/\lambda$. For
$\lambda=1$, every node with nonzero weight has to contain all $n$
samples and only the root node is thus selected for $\lambda=1$. At the
other extreme, let $m$ be the minimal node size (here $m=5$). For
$\lambda> n/m$, the constraint will have no effect at all, since all
nodes have $n_g\ge m$ anyhow and the average weighted fraction
(\ref{weightedfraction}) is thus bounded from below by $m/n$ for all
weight vectors. The regularized estimator $\hat{\w}^\lambda$ solves
then
%
\begin{eqnarray}\label{eq:whatlambda}
\hat{\w}^\lambda=\operatorname{argmin}\limits_{\w}\|\Y-\M\w\|^2\nonumber
\\[-9.5pt]\\[-9.5pt]
\eqntext{\mbox{such that }\I\w=1\mbox{ and }\w\ge0\mbox{ and }\|\w\|_1\le\lambda}
\end{eqnarray}
instead of
(\ref{eq:what}). The interesting region is $\lambda\in[1,m/n]$, where
$m$ is the enforced lower bound on node size. From the point of
predictive accuracy, constraining $\lambda$ is usually not beneficial
unless the signal-to-noise ratio is very low. There is thus a tradeoff
between sparsity (number of selected nodes) and predictive power, as
shown in the next section with numerical results.

\section{Numerical results}
\label{section:numerical}

For various data sets, we look at the predictive accuracy of
\textit{node harvest} (NH) and various related aspects like sensitivity
to the size of the initial set of nodes, interpretability and
predictive power of results under additional regularization as in
(\ref{eq:whatlambda}).

\subsection{Example I: Two-dimensional sinusoidal reconstruction}

\begin{figure}
\begin{tabular}{@{}cc@{}}
(a)&(b)\\

\includegraphics{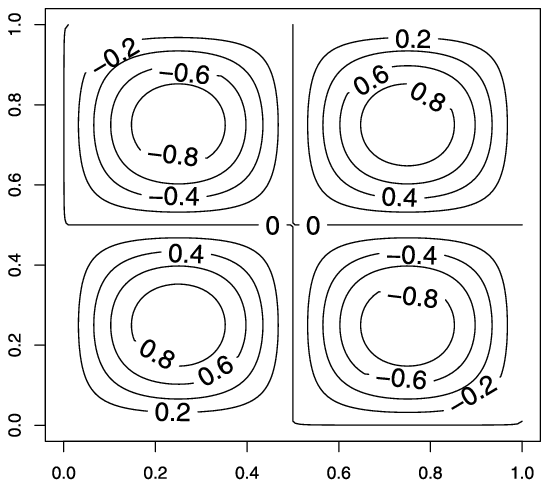}
&\includegraphics{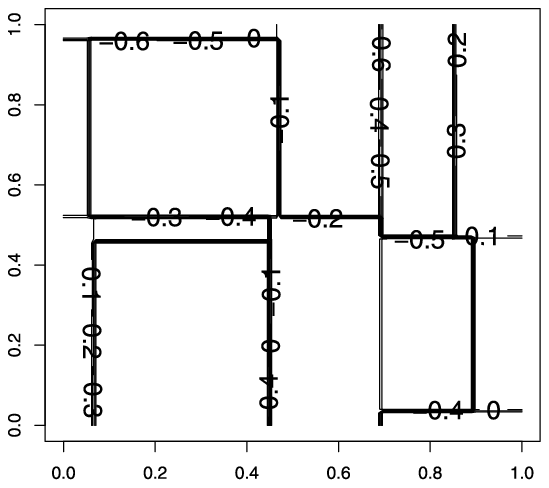}\\
(c)&(d)\\

\includegraphics{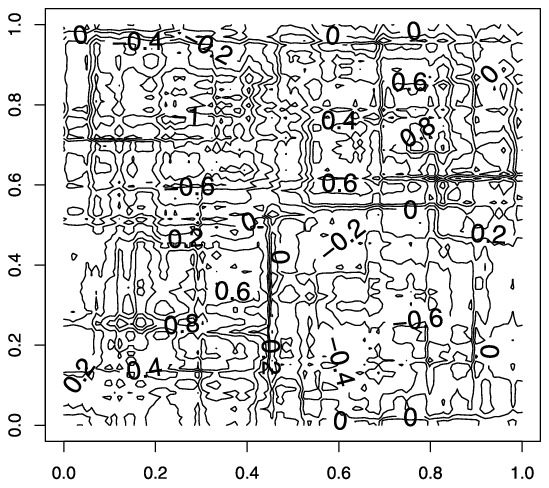}
&\includegraphics{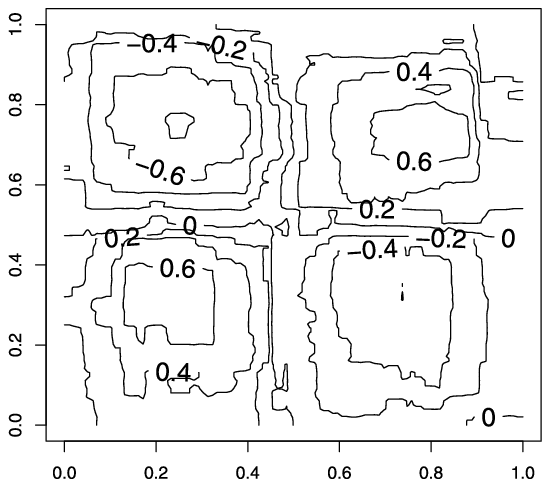}\\
\end{tabular}
\caption{\textup{(a)} Contour plot of $E(Y)$ under model (\protect\ref{Ytoy}) in the
two-dimensional predictor space, with contour lines at values $-$1 to 1
with step sizes of 0.2. The contour plot for the fit of a regression
tree~\textup{(b)}, a Random Forest fit \textup{(c)} and
\textit{node harvest} \textup{(d)}. The three methods are fitted using the
same $10^3$ observations from (\protect\ref{Ytoy}).}\label{fig:sine}
\end{figure}

As a very simple first example, assume that the random predictor
variable $\mathbf{x}=(\mathbf{x}_1,\mathbf{x}_2)$ is two-dimensional
and distributed uniformly
on $[0,1]^2$. and the response is generated as
%
\begin{equation}\label{Ytoy}
Y=\sin(2\pi\mathbf{x}_1)\sin(2\pi\mathbf{x}_2)+\varepsilon,
\end{equation}
where $\varepsilon$ follows a normal distribution with mean 0 and
variance $1/4$ and the noise is independent between observations.
Taking $n=10^3$ samples from (\ref{Ytoy}), a regression tree
[\citet{CART}] is fitted to the data, using a cross-validated choice of
tree size penalty. The fit is constant on rectangular regions of the
two-dimensional space, as shown in Figure \ref{fig:sine}. Each of these
regions corresponds to a node in the tree. The fit is rather poor,
however, and the structure of the problem is not well captured. Random
Forests is fitted with the default parameters proposed in
\citet{breiman01random}.
It improves in terms of predictive accuracy on
trees, yet the contour plot appears very noisy since the trees are
grown until almost pure (keeping only 10 observations in each node) and
the variability of the Random Forests approach manifests itself here
in a high spatial variability of the fitted function. NH is fitted with
the default parameters used throughout (1000 random nodes generated
picked from a Random Forest fit, two-factor interactions and minimal
node size of 10). It gives a comparably clean contour plot, as seen in
the rightmost panel of Figure \ref{fig:sine} and forms a compromise
between trees and Random Forests. In contrast to trees, the fitted
function is not constant across rectangular-shaped subspaces since
each observation can fall into more than one node.

\subsection{Example II: Importance sampling in climate modeling}

The \textit{climate\-prediction.net} project [\citet{allen1999yourself}]
is, broadly speaking, concerned with uncertainty analysis of climate
models, using a distributed computing environment. A climate model
contains typically several parameters whose precise values are only
known up to a certain precision. The project analyzes the behavior of a
coarse resolution variant of the HadCM3 climate model
[\citet{johns2003anthropogenic}] under thousands of small perturbations
of the default parameters. Once a certain number of models has been
sampled, the behavior of the underlying climate model can be better
understood and importance sampling can be used to sample only in
relevant sections of the parameter space. While Gaussian process
emulation is widely used in this context
[\citet{oakley2004probabilistic}], we note that the data here are not
noise free since the outcome depends on the random initial conditions
and a standard regression analysis of the model is hence useful.
Without giving a full explanation, we show an example of a data set
containing 250 models, each run with a different combination of 29
parameters. The response variable is mean temperature change over a 50
year period under a given emissions scenario.

\begin{figure}

\includegraphics{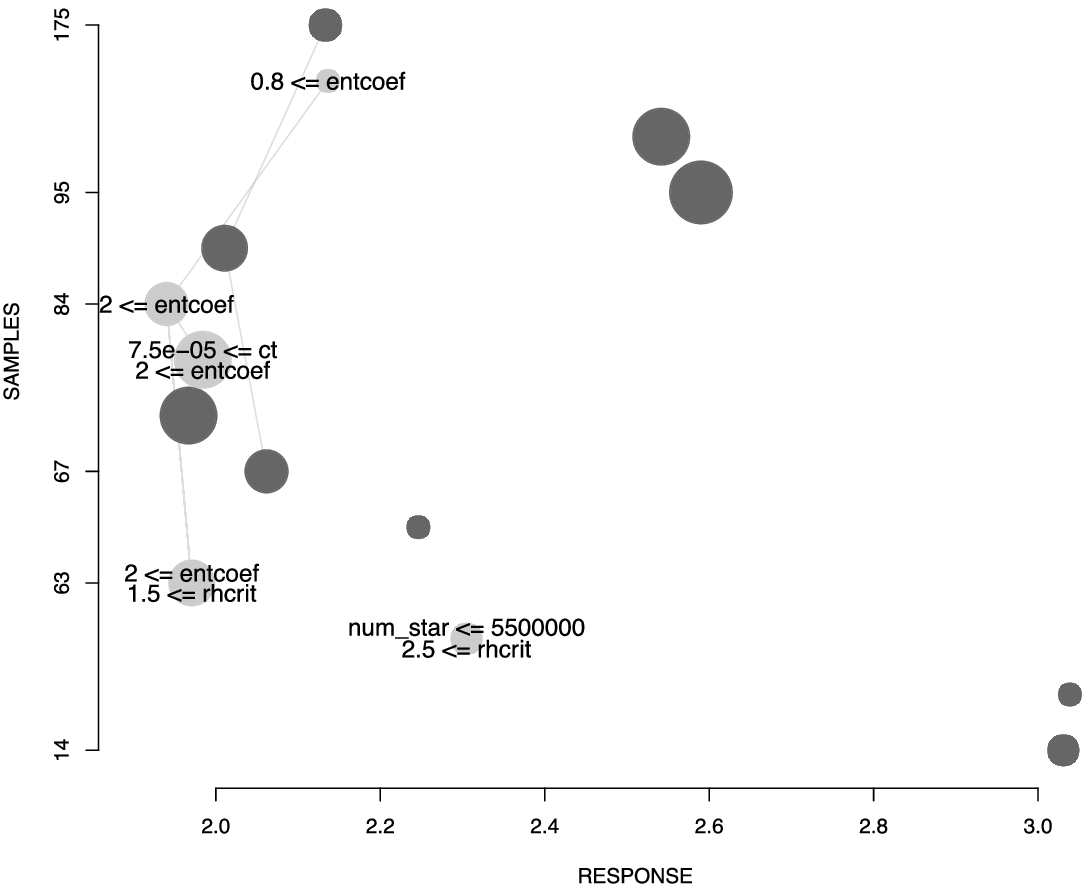}

\caption{The 14 nodes selected by \textit{node harvest} for the
\textup{climateprediction.net} data. The area of each node $g$ is
proportional to the weight $\hat{\w}_g$ it received in
(\protect\ref{eq:what}). The 4986 nodes that received a zero weight are not
shown. The position on the x-axis shows for each node $g$ the mean
$\mu_g$ of all training observations that fall into it, while the
position on the y-axis shows how many observations it contains. If
observations of a node are a subset of observations of another node, a
line between the two nodes is drawn. The node ``$ \mathit{entcoef}\ge
2$'' contains a subset of the observations of the node ``$
\mathit{entcoef}\ge0.8$.'' A single new observation was chosen at
random and the 5 nodes that the new observation falls into are lighter
and annotated. The prediction for the new observation is then simply
the weighted mean across the x-axis positions of the annotated nodes.}\label{fig:cpdn}
\end{figure}

Following the approach laid out above, 1000 nodes are generated with a
Random Forest type approach. All of these nodes are constrained to
contain at least 10 observations and have at most two-factor
interactions. Then the quadratic program~(\ref{eq:what}) is applied.
Only 14 of the originally 1000 nodes receive a nonzero weight and these
nodes are shown in Figure \ref{fig:cpdn}.

The plot is very interpretable: the position of each node on the $x$-axis
corresponds to the mean of all training observations in this node. And
predictions for new data are simply the weighted mean across all nodes
the new observation falls into. The weight of each node is proportional
to the area with which it is plotted.

To give an example of a prediction, a new observation is sampled at
random. It happens to fall into five nodes, whose respective weights
and node means are as follows:

\begin{center}
\tabcolsep=3pt
\fontsize{8.5}{11.5}\selectfont{
\begin{tabular*}{\tablewidth}{@{\extracolsep{\fill}}lccccc@{}}
\hline
&$\bolds{\mathit{entcoef}\ge2}$
&$\bolds{\mathit{entcoef} \ge2}$&
&$\bolds{\mathit{num\_star} \le5.5\cdot10^5}$&\\
\textbf{Node} $\bolds{g}$&$\bolds{\mathit{ct}\ge7.5\cdot10^{-5}}$&$\bolds{\mathit{rhcrit}\ge1.5}$&$\bolds{\mathit{entcoef} \ge2}$&$\bolds{\mathit{rhcrit}\ge2.5}$&$\bolds{\mathit{entcoef} \ge.8 }$\\
\hline
Mean $\mu_g$ &1.98&1.97&1.94&2.30&2.14 \\
Weight $\hat{\w}_g$ &0.37&0.24&0.21&0.11&0.06\\
\hline
\end{tabular*}}
\end{center}

\noindent
Four of these nodes contain the entrainment coefficient
($\mathit{entcoef}$) as a split variable, which is maybe unsurprising
since the entrainment coefficient is known to be the parameter to
which the model is most sensitive.

The new observation belongs also to the root node (as do all
observations), with the minimal imposed weight $0{.}001$ for this node,
but this influence is negligible and ignored here. The predicted
response for this new observation is then the weighted mean across
these nodes, which is $ 2.014$. A graphical visualization of this
weighted averaging is immediate from Figure \ref{fig:cpdn}. The
prediction for this new observation (or rather model) is simply the
weighted horizontal position of the 5 selected and annotated nodes,
with weights proportional to node size. As will be seen further below,
the predictive accuracy of NH is for this data set better than
cross-validated trees, even though no tuning was used in the NH
approach and the result is at least as interpretable and simple as a
tree. To get optimal predictive performance, a tree needs to employ
interactions up to fourth order while NH gets a better accuracy with
only two-factor interactions.

\subsection{Example III: Wisconsin breast cancer data}

\begin{figure}

\includegraphics{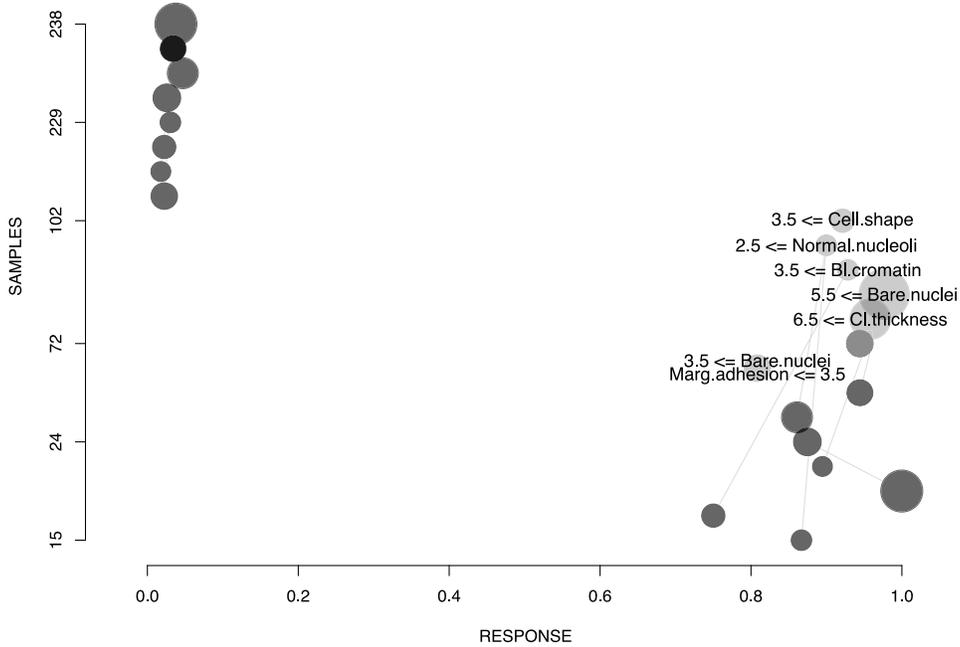}

\caption{\textit{Node harvest} (NH) estimator for the Wisconsin Breast
Cancer study. 22 nodes are selected, where the number of patients
within each node is shown on the vertical scale. The percentage of
patients with a malignant tumor ($Y=1$) is shown for each node on the
horizontal scale. The number of patients within each node is shown on
the vertical scale. The size of nodes is again plotted proportional to
the weights chosen by NH. A new patient was randomly selected and
belongs to the 6 lighter and annotated nodes. Among these, there are 5
`main effect' nodes, with the addition of one `two-factor interaction'
node. All of the 6 selected groups of patients contain a large fraction
of people with a malignant tumor, with actual proportions varying
between 83\% for node ``$\mathit{Bare.nuclei} \ge3.5 ;
\mathit{Marg.adhesion} \le3.5$'' to above 97\% for node
``$\mathit{bare.nuclei} \ge5.5.$'' The estimated probability for
having a malignant tumor for this new patient is the weighted mean
across the percentages of people with a malignant tumor in these 6
groups of patients.}\label{fig:breast}
\end{figure}

As an example of binary classification, take the Wisconsin breast
cancer data [\citet{mangasarian1995breast}]. There are 10 clinical
variables to predict whether a tumor is benign or malignant. Applying
NH again with 1000 RF-generated nodes, with at most two-factor
interactions and a minimal node size of 10, the results in
Figure~\ref{fig:breast} are obtained. The root node is again not
shown, despite its small enforced positive weight of $0{.}001$. The
position on the $x$-axis gives for each node the percentage of people
within this group that had a malignant tumor ($Y=1$). The $y$-axis
position is proportional to the number of people within this node in
the training sample. A new patient falls into one or several of these
nodes and the predicted probability of class $Y=1$ for this patient is
simply the weighted average over the means $\mu_g$ of all nodes $g$ the
patient is part of, as shown for a randomly chosen example patient in
Figure~\ref{fig:breast}. A prediction (or risk assessment in the
example) is thus easy to communicate and can be related to the
empirical outcome in relevant groups of patients with similar
characteristics.

If splitting the data into two equally large parts and taking one part
as training and the other part as test data, and averaging over 20
splits, the misclassification test error with NH is 3.6\%, compared
with 3.3\% for Random Forests and 5.5\% for cross-validated
classification trees. NH seems to perform better in a low
signal-to-noise ratio setting. If changing 20\% of all labels in the
training set, the performance of Random Forests drops to 6.0\% while NH
maintains an accuracy of 4.4\%. This behavior is completely analogous
to regression, as shown below.

\subsection{Further data sets}
Besides these examples, the method is applied to motif regression
[\citet{conlon03motif}], where the task is to identify transcription
factor binding sites from gene expression measurements. The data set
consist of $n=2588$ samples and $p=660$ genes and the response variable
is the concentration of the transcription factor. In addition, the
well-known abalone data [\citet{abalone}], with $p=8$, are considered,
as are the diabetes data from \citet{efron04least} (`diabetes,' $p=10,
n=442$) and the LA Ozone data (`ozone,' $p=9, n=203$), bone mineral
density data (`bones,' $p=4, n=485$), fuel efficiency data (`mpg,'
$p=7, n=392$), median house prices in the Boston area (`housing,'
$p=13, n=506$), CPU performance data (`machine,' $p=7, n=209$), crime
rate data from the US census (`crime,' $p=101, n=1993$), and a data set
about prediction of Parkinson's symptoms from voice measurements
(`parkinson,' $p=19, n=5875$). The latter data sets are all available
at the UCI machine learning repository [\citet{UCI}]. We also consider a
data set about radial velocity of galaxies (`galaxy,' $p=4, n=323$) and
prostate cancer analysis (`prostate,' $p=8, n=97$), the latter all from
\citet{hastie2001esl}, which contains more details, and, finally, a
gene expression data set, kindly provided by DSM nutritional products
(Switzerland). For $n = 115$ samples, there is a continuous response
variable measuring the logarithm of riboavin (vitamin B2) production
rate of Bacillus Subtilis, and there are $p = 4088$ continuous
covariates measuring the logarithm of gene expressions from essentially
the whole genome of Bacillus Subtilis. Certain mutations of genes are
thought to lead to higher vitamin concentrations and the challenge is
to identify those relevant genes via regression, possibly using also
interaction between genes. Observations with missing values are removed
from the data sets. Even though NH could deal with these, as alluded to
above, it facilitates comparison with other techniques.

Each data set is split 10 times into two equally large parts. On the
half used as a training set, NH is employed as well as Random Forests
(RF), a CART regression tree (TREE), Rule Ensembles (RE) and
$L_2$-boosted regression trees ($L_2$B). For NH, we select 1000 nodes
from the Random Forest ensemble as described above, keeping only
main-effect and two-factor interaction nodes and a minimal node size of
5. Then (\ref{eq:what}) is applied to this ensemble and exactly the
same procedure is followed for all data sets without any tuning of
these parameters. The same initial set of nodes is used for Rule
Ensembles with a 5-fold CV-choice of the tuning parameter. We remark
that both NH and RE could perform better for some data sets if higher
order interactions were allowed in the nodes. For Random Forests, one
could fine tune the minimal node size or the value of \textit{mtry},
which is the size of the random number of variables used to find the
optimal split point at each node. However, they are kept at the default
values (which are known to give nearly optimal results), as proposed in
\citet{breiman01random} and \citet{liawclassification}, to give an
equal comparison between the two essentially `tuning'-free algorithms
NH and RF. The size of the regression trees [\citet{CART}] is chosen by
10-fold CV on the training data. Boosting is using regression trees of
depth two as weak learners and a CV-optimized stopping time. The
predictions on the test data (the second part of the data) are then
recorded for all three methods and the fraction of the variance that is
unexplained is averaged across all 10 sample splits. The number of
training observations available for each data set is shown in Table~\ref{table:1}, together with the average unexplained fraction of the
variance.

\begin{table}
\tabcolsep=0pt
\caption{Average proportion of unexplained variance on
test data, rounded to two significant figures for Random Forests (RF),
CART trees (TREE), Node Harvest without regularization,
$\lambda=\infty$, ($\textrm{NH}_{\infty}$), Rule Ensembles (RE) and
$L_2$-boosted regression trees ($L_2$B)}\label{table:1}
\begin{tabular*}{\textwidth}{@{\extracolsep{\fill}}lcc@{\hspace*{3pt}}ccccc@{\hspace*{3pt}}ccccc@{}}
\hline
&&&&&&&&\multicolumn{5}{c@{}}{\textbf{With additional}}\\
&&&&&&&&\multicolumn{5}{c@{}}{\textbf{observational noise}}\\[-5pt]
&&&&&&&&\multicolumn{5}{c@{}}{\hrulefill}\\
\textbf{Data set}&$\bolds{n}$&$\bolds{p}$&\textbf{RF}&\textbf{TREE}&\textbf{RE}&$\bolds{L_2}$\textbf{B}&\textbf{NH}$_{\bolds{\infty}}$
&\textbf{RF}&\textbf{TREE}&\textbf{RE}&$\bolds{L_2}$\textbf{B}&\textbf{NH}$_{\bolds{\infty}}$\\
\hline
Ozone&\phantom{0}203&\phantom{00}12&\textbf{0.27}&\textit{0.41}&0.31&0.33&0.34&0.55&$>$1&$>$1&0.67&\textbf{0.47}\\
Mpg&\phantom{0}392&\phantom{000}7&\textbf{0.15}&\textit{0.24}&0.16&0.16&0.20&0.54&$>$1&0.47&0.39&\textbf{0.36}\\
Servo&\phantom{0}166&\phantom{000}4&0.32&0.38&\textbf{0.20}&0.37&0.26&0.61&\textit{0.94}&0.73&0.88&\textbf{0.57}\\
Prostate&\phantom{00}97&\phantom{000}8&\textbf{0.53}&\textit{0.68}&0.61&0.63&0.58&$>$1&$>$1&$>$1&$>$1&$>$1\\
Housing&\phantom{0}506&\phantom{00}13&\textbf{0.14}&\textit{0.30}&0.18&0.19&0.25&0.46&$>$1&0.66&0.48&\textbf{0.39}\\
Diabetes&\phantom{0}442&\phantom{00}10&\textbf{0.55}&\textit{0.71}&0.58&0.57&0.59&0.74&$>$1&$>$1&0.74&\textbf{0.65}\\
Machine&\phantom{0}209&\phantom{000}7&\textbf{0.16}&0.58&\textit{0.86}&0.43&0.27&0.84&$>$1&$>$1&0.56&\textbf{0.54}\\
Galaxy&\phantom{0}323&\phantom{000}4&\textbf{\phantom{0}0.036}&\textit{\phantom{0}0.094}&\phantom{0}0.045&\phantom{0}0.049&\phantom{0}0.065&0.53&\textit{0.81}&0.35&0.33&\textbf{0.26}\\
Abalone&4177&\phantom{000}8&\textbf{0.45}&0.56&0.52&0.48&\textit{0.60}&0.64&\textit{0.65}&0.59&\textbf{0.56}&0.61\\
Bones&\phantom{0}485&\phantom{000}3&0.71&\textit{0.79}&0.73&0.73&\textbf{0.70}&\textbf{0.83}&$>$1&0.98&0.88&0.85\\
Cpdn&\phantom{0}493&\phantom{00}29&\textbf{0.52}&0.66&0.55&\textit{0.68}&0.66&0.98&$>$1&0.98&0.98&\textbf{0.77}\\
Motifs&2587&\phantom{0}666&\textbf{0.67}&\textit{0.87}&0.72&0.71&0.78&0.83&$>$1&$>$1&0.84&\textbf{0.80}\\
Vitamin&\phantom{0}115&4088&\textbf{0.35}&\textit{0.55}&0.40&0.38&0.37&\textbf{0.78}&$>$1&$>$1&0.99&0.98\\
Crime&1993&\phantom{0}101&\textbf{0.34}&\textit{0.47}&0.38&0.36&0.42&0.46&\textit{0.70}&0.49&0.46&\textbf{0.45}\\
Parkinson&5875&\phantom{00}19&\textbf{0.20}&0.33&0.53&0.63&\textit{0.69}&\textbf{0.43}&0.68&0.60&\textit{0.76}&0.69\\
\hline
\end{tabular*}
\tabnotetext[]{}{\textit{Notes}: The best performing method is
shown in bold, while the worst performing method
is shown in italics. A result `$>$1' indicates
that the prediction is worse on test data than the best constant
prediction.}
\end{table}

\begin{figure}

\includegraphics{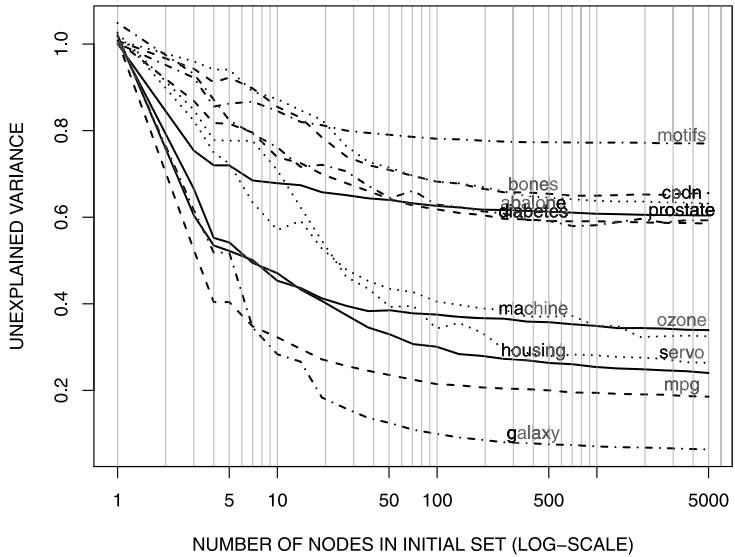}

\caption{The unexplained variance on test data as a function of the
number $q$ of nodes in the initial set of nodes (x-axis in log-scale).
Each line corresponds to one data set. Close to optimal performance is
reached after a few hundred nodes, with results continuing to improve
slightly thereafter.}\label{fig:nonodes}
\end{figure}

On most data sets, Random Forests has the highest predictive accuracy
with the exception of `servo' and `bones,' where NH is coming on top. A
single tree is, maybe unsurprisingly, consistently the worst performing
method. The picture changes if additional noise is added to the
training observations. To this end, the response vector $\Y$ is
replaced on the training observations with the response $\Y+
\varepsilon$, where $\varepsilon=(\varepsilon_1,\ldots,\varepsilon_n)$
contains i.i.d. standard normal noise with variance three times the
variance of $\Y$, cutting the correlation between the true unknown
signal and the response exactly in half. As can be seen in the right
part of the table, NH is now the best performing method on the clear
majority of these low signal-to-noise ratio data, sometimes
outperforming all other approaches by a substantial margin.

Figure \ref{fig:nonodes} shows the impact that the number of nodes in
the initial set $\mathcal{Q}$ has on predictive accuracy: the more
nodes in $\mathcal{Q}$, the better the predictive accuracy on test
data. Even though Figure \ref{fig:nonodes} shows this phenomenon only
up to a few thousands of nodes, it holds well beyond this point. In
other words, NH does not seem to overfit if more and more nodes are
added to the initial set of nodes and it is ideal to include as many
nodes as computationally feasible in $\mathcal{Q}$, even though a few
hundred seem to be sufficient for most data sets.

A crude measure for the complexity of a tree or tree ensembles is the
total number of nodes of the predictor, which is equivalent to the
total number of leaf nodes for tree ensembles and the total number of
nodes with nonzero weights or coefficients for NH and RE respectively.
Table~\ref{table:nodescomp} shows that NH (with $\lambda=\infty$) and
RE use roughly a similar amount of nodes in the final fit, typically a
few dozen, while NH with regularization yields the sparsest results in
general, with the obvious exception of single trees, as seen in the
following Table~\ref{table:2}. Boosting leads to hundreds and Random
Forests to thousands or even hundreds of thousands of final leaf nodes.
The greater sparsity of NH and RE comes at a higher computational
price. Starting from the same number of initial nodes, NE and RE are
more computationally intensive to compute than all other methods, with
a slight edge for NH, especially for data sets with a larger sample
size. While it is faster to fit RF than either RE or NH, it should be
emphasized that, due to much fewer used nodes, NH and RE are clearly
very fast for predicting the response of new observations, which can be
of importance in an online prediction setting, where RF can be too slow
for some applications.

\begin{table}
\tabcolsep=1pt
\caption{\label{table:nodescomp} Average number of nodes for each
tree-based predictor (left half) and the average computational time
necessary to fit the predictor in seconds (right half), rounded to two
significant figures}
\begin{tabular*}{\textwidth}{@{\extracolsep{\fill}}ld{4.1}d{4.1}@{\hspace*{3pt}}cd{2.1}cd{3.1}c@{\hspace*{3pt}}d{3.3}d{1.4}d{3.1}d{3.2}d{2.1}@{}}
\hline
&&&\multicolumn{5}{c@{\hspace*{3pt}}}{\textbf{Number of leaf nodes}}&\multicolumn{5}{c@{}}{\textbf{Computational time (s)}}\\[-5pt]
&&&\multicolumn{5}{c@{\hspace*{3pt}}}{\hrulefill}&\multicolumn{5}{c@{}}{\hrulefill}\\
\textbf{Data set}&\multicolumn{1}{c}{$\bolds{n}$}&\multicolumn{1}{c}{$\bolds{p}$}&\textbf{RF}&\multicolumn{1}{c}{\textbf{TREE}}&\textbf{RE}
&\multicolumn{1}{c}{$\bolds{L_2}$\textbf{B}}&\textbf{NH}$_{\bolds{\infty}}$&\multicolumn{1}{c}{\textbf{RF}}&\multicolumn{1}{c}{\textbf{TREE}}
&\multicolumn{1}{c}{\textbf{RE}}&\multicolumn{1}{c}{$\bolds{L_2}$\textbf{B}}&\multicolumn{1}{c@{}}{\textbf{NH}$_{\bolds{\infty}}$}\\
\hline
Ozone&203&12&$>$$10^4$&6.7&32&98&97&0.15&0.016&11&0.57&25\\
Mpg&392&7&$>$$10^4$&8&38&150&56&0.17&0.015&28&0.53&8.6\\
Servo&166&4&$>$$10^4$&2.9&21&110&20&0.074&0.0098&3.8&0.27&3.2\\
Prostate&97&8&$>$$10^4$&3.9&20&71&52&0.22&0.01&2.8&0.27&8.8\\
Housing&506&13&$>$$10^4$&8.3&60&130&74&0.36&0.025&84&0.53&24\\
Diabetes&442&10&$>$$10^4$&12&36&96&75&0.26&0.019&57&0.5&31\\
Machine&209&7&$>$$10^4$&3.6&62&420&47&0.24&0.011&9&0.34&6.7\\
Galaxy&323&4&$>$$10^4$&5.8&49&170&52&0.19&0.0098&19&0.35&8.6\\
Abalone&4177&8&$>$$10^4$&11&74&150&53&5.8&0.16&520&0.76&38\\
Bones&485&3&$>$$10^4$&10&27&67&30&0.15&0.011&20&0.45&4.6\\
Cpdn&493&29&$>$$10^4$&13&44&82&24&0.42&0.041&35&0.71&4.8\\
Motifs&2587&666&$>$$10^4$&15&68&120&64&140&11&470&100&86\\
Vitamin&115&4088&$>$$10^4$&4.7&43&230&70&2.5&0.92&4.6&170&75\\
Crime&1993&101&$>$$10^4$&11&59&140&71&12&0.83&220&3&21\\
Parkinson&5875&19&$>$$10^4$&24&97&100&15&16&0.53&920&1.4&44\\
\hline
\end{tabular*}
\end{table}

\begin{table}
\tabcolsep=0pt
\caption{Average proportion of unexplained variance
and average number of selected nodes for the unrestricted \textit{node
harvest} estimator ($\lambda=\infty$) and the regularized estimator
($\lambda=3$), where the average fraction of samples in each node has
to be larger than $\lambda^{-1}=1/3$. The better performing method is
again shown in bold}\label{table:2}
\begin{tabular*}{\textwidth}{@{\extracolsep{4in minus 4in}}l@{\hspace*{6pt}}ccc@{\hspace*{3pt}}ccc@{\hspace*{6pt}}ccc@{\hspace*{3pt}}cc@{\hspace*{3pt}}c@{}}
\hline
&&&&&&&\multicolumn{6}{c@{}}{\textbf{With additional noise}}\\
&&&&&&&\multicolumn{6}{c@{}}{\hrulefill}\\
&\multicolumn{3}{c@{\hspace*{3pt}}}{\textbf{Unexpl. variance}}&\multicolumn{3}{c@{\hspace*{6pt}}}{\textbf{No. selected nodes}}
&\multicolumn{3}{c@{\hspace*{3pt}}}{\textbf{Unexpl. variance}}
&\multicolumn{3}{c@{}}{\textbf{No. selected nodes}}\\[-5pt]
&\multicolumn{3}{c@{\hspace*{3pt}}}{\hrulefill}&\multicolumn{3}{c@{\hspace*{6pt}}}{\hrulefill}&\multicolumn{3}{c@{\hspace*{3pt}}}{\hrulefill}
&\multicolumn{3}{c@{}}{\hrulefill}\\
\textbf{Data set}&$\bolds{\lambda=\infty}$&&\multicolumn{1}{c@{\hspace*{3pt}}}{$\bolds{\lambda=3}$}&$\bolds{\lambda=\infty}$
&&$\bolds{\lambda=3}$&$\bolds{\lambda=\infty}$&&$\bolds{\lambda=3}$&$\bolds{\lambda=\infty}$&&$\bolds{\lambda=3}$\\
\hline
Ozone&\textbf{0.34}&&0.34&97&&\textbf{73}&\textbf{0.47}&&0.48&100&&\textbf{95}\\
Mpg&\textbf{0.20}&&0.24&56&&\textbf{37}&0.36&&\textbf{0.34}&35&&\textbf{34}\\
Servo&\textbf{0.26}&&0.27&20&&\textbf{11}&0.57&&\textbf{0.49}&23&&\textbf{17}\\
Prostate&0.58&&\textbf{0.57}&52&&\textbf{47}&$>$1&&\textbf{0.98}&53&&\textbf{47}\\
Housing&\textbf{0.25}&&0.28&74&&\textbf{40}&\textbf{0.39}&&0.41&69&&\textbf{46}\\
Diabetes&\textbf{0.59}&&0.61&75&&\textbf{55}&\textbf{0.65}&&0.66&96&&\textbf{96}\\
Machine&0.27&&\textbf{0.26}&47&&\textbf{35}&0.54&&\textbf{0.48}&42&&\textbf{40}\\
Galaxy&\phantom{0}\textbf{0.065}&&\phantom{0}0.097&52&&\textbf{32}&0.26&&\textbf{0.24}&44&&\textbf{33}\\
Abalone&\textbf{0.60}&&0.63&53&&\textbf{38}&\textbf{0.61}&&0.63&39&&\textbf{28}\\
Bones&0.70&&\textbf{0.70}&30&&\textbf{22}&0.85&&\textbf{0.83}&30&&\textbf{24}\\
Cpdn&\textbf{0.66}&&0.68&24&&\textbf{18}&\textbf{0.77}&&0.79&48&&\textbf{36}\\
Motifs&\textbf{0.78}&&0.78&64&&\textbf{46}&0.80&&\textbf{0.80}&54&&\textbf{42}\\
Vitamin&\textbf{0.37}&&0.39&70&&\textbf{60}&1.00&&\textbf{0.85}&71&&\textbf{66}\\
Crime&\textbf{0.42}&&0.44&71&&\textbf{44}&\textbf{0.45}&&0.48&59&&\textbf{46}\\
Parkinson&\textbf{0.69}&&0.71&15&&\textbf{12}&\textbf{0.69}&&0.73&22&&\textbf{18}\\
\hline
\end{tabular*}
\end{table}

Last, the effect of regularization (\ref{eq:whatlambda}) on the
sparsity of the solution and predictive accuracy is examined. Results
are summarized in Table~\ref{table:2}, where the unconstrained
estimator is compared for all previous data sets with the regularized
estimator at $\lambda=3$. Unsurprisingly, regularization always
improves the sparsity of the solution. The average number of selected
nodes can decrease by a potentially substantial amount if applying the
additional regularization, improving interpretability. Predictive
accuracy is typically very similar between the two estimators, with an
advantage for the unconstrained estimator for the original data sets.
Regularization seems to improve the already very good performance of NH
in the low signal-to-noise ratio setting where additional noise is
applied to the training data. Overall, the unconstrained estimator
seems a very good default choice. Applying the additional
regularization is worthwhile if the results are desired to be very
sparse or the signal in the data is extremely weak.

\section{Discussion}

The aim of \textit{node harvest} (NH) is to combine positive aspects of
trees on the one hand and tree ensembles such as Random Forests on the
other hand.

NH shares with trees the ease of interpretability and simplicity of
results. As with trees, only a few nodes are used. For trees, every
observation falls exactly into one such node and the predicted response
is the corresponding node mean. With NH, nodes can overlap and an
observation can be a member of a few nodes. While trees often have to
include higher order interactions to achieve their optimal predictive
performance, it is often sufficient for NH to include main effects and
two-factor interactions. While tree size is determined by
cross-validation, essentially no tuning parameter and no
cross-validation is necessary for NH.

The lack of a very important tuning parameter is thus a common feature
of both NH and Random Forests. Predictive accuracy also seems
comparable. For high signal-to-noise ratio data, Random Forests seems
to have an edge while NH delivers typically a smaller loss if the
signal-to-noise ratio drops to lower values. The general advantage of
NH over Random Forests is simplicity and arguably much better
interpretability of results.

In common with both trees and tree ensembles, NH can handle mixed data
very well and is invariant under monotone transformations of the data.
NH is, moreover, able to deal with missing values without explicit use
of imputation or surrogate splits. Both regression and classification
are handled naturally and it is conceivable that the method can also be
extended to censored data, in particular, survival analysis, in analogy
to the extension of Random Forests to Random Survival Forests
[\citet{ishwaran2006random}]. Most of the functionality of \textit{node
harvest} is implemented in the package \texttt{nodeHarvest} for the
\texttt{R}-programming language [\citet{R}].

\section*{Acknowledgments}
I would like to thank a referee, an Associate
Editor and the editor Michael Stein for their helpful comments on an
earlier version of the manuscript.


\printaddresses


\begin{thebibliography}{99}

\bibitem[\protect\citeauthoryear{Allen}{1999}]{allen1999yourself}
\textsc{Allen, M.} (1999). Do-it-yourself climate prediction.
\textit{Nature} \textbf{401} 642--642.

\bibitem[\protect\citeauthoryear{Amit and Geman}{1997}]{amit97shape}
\textsc{Amit, Y.} and \textsc{Geman, D.} (1997). Shape quantization and
recognition with randomized trees. \textit{Neural Comput.} \textbf{9}
1545--1588.

\bibitem[\protect\citeauthoryear{Asuncion and Newman}{2007}]{UCI}
\textsc{Asuncion, A.} and \textsc{Newman, D.} (2007). \textit{UCI
Machine Learning Repository}. Univ. California, Irvine, CA.

\bibitem[\protect\citeauthoryear{Bartlett, Jordan and
McAuliffe}{2003}]{bartlett03convexity}
\textsc{Bartlett, P., Jordan, M.} and \textsc{McAuliffe, J.} (2003).
Convexity, classification, and risk bounds. Technical report,
Dept. Statistics, UC Berkeley.

\bibitem[\protect\citeauthoryear{Blanchard et~al.}{2007}]{blanchard2007optimal}
\textsc{Blanchard, G., Sch{\"a}fer, C., Rozenholc, Y.} and
\textsc{M{\"u}ller, K.} (2007). Optimal dyadic decision trees.
\textit{Mach. Learn.} \textbf{66} 209--241.

\bibitem[\protect\citeauthoryear{Breiman}{1996a}]{breiman1996bagging}
\textsc{Breiman, L.} (1996a). Bagging predictors. \textit{Mach. Learn.}
\textbf{24} 123--140.

\bibitem[\protect\citeauthoryear{Breiman}{1996b}]{breiman1996stacked}
\textsc{Breiman, L.} (1996b). Stacked regressions. \textit{Mach.
Learn.} \textbf{24} 49--64.

\bibitem[\protect\citeauthoryear{Breiman}{2001}]{breiman01random}
\textsc{Breiman, L.} (2001). Random forests. \textit{Mach. Learn.}
\textbf{45} 5--32.

\bibitem[\protect\citeauthoryear{Breiman et~al.}{1984}]{CART}
\textsc{Breiman, L.,~Friedman, J., Olshen, R.} and \textsc{Stone, C.}
(1984). \textit{Classification and Regression Trees}. Wadsworth,
Belmont, CA.
\MR{0726392}

\bibitem[\protect\citeauthoryear{Chen, Donoho and
Saunders}{2001}]{chen01atomic}
\textsc{Chen, S., Donoho, S.} and \textsc{Saunders, M.} (2001). Atomic
decomposition by basis pursuit. \textit{SIAM Rev.} \textbf{43}
129--159.
\MR{1854649}

\bibitem[\protect\citeauthoryear{Conlon et~al.}{2003}]{conlon03motif}
\textsc{Conlon, E., Liu, X., Lieb, J.} and \textsc{Liu, J.} (2003).
Integrating regulatory motif discovery and genome-wide expression
analysis. \textit{Proc. Natl. Acad. Sci.} \textbf{100} 3339--3344.

\bibitem[\protect\citeauthoryear{Efron et~al.}{2004}]{efron04least}
\textsc{Efron, B., Hastie, T., Johnstone, I.} and
\textsc{Tibshirani,~R.} (2004). Least angle regression. \textit{Ann.
Statist.} \textbf{32} 407--451.
\MR{2060166}

\bibitem[\protect\citeauthoryear{Freund and Schapire}{1996}]{freund1996enb}
\textsc{Freund, Y.} and \textsc{Schapire, R.} (1996). Experiments with
a new boosting algorithm. In \textit{Machine Learning: Proceedings of
the Thirteenth International Conference} 148--156. Morgan Kauffman,
San Francisko, CA.

\bibitem[\protect\citeauthoryear{Friedman, Hastie and
Tibshirani}{2000}]{friedman00additive}
\textsc{Friedman, J., Hastie, T.} and \textsc{Tibshirani, R.} (2000).
Additive logistic regression: A statistical view of boosting.
\textit{Ann. Statist.} \textbf{28} 337--407.
\MR{1790002}

\bibitem[\protect\citeauthoryear{Friedman and
Popescu}{2008}]{friedman2005plv}
\textsc{Friedman, J.} and \textsc{Popescu, B.} (2008). Predictive
learning via rule ensembles. \textit{Ann. Appl. Statist.} \textbf{2}
916--954.
\MR{2522175}

\bibitem[\protect\citeauthoryear{Goldfarb and
Idnani}{1983}]{goldfarb1983numerically}
\textsc{Goldfarb, D.} and \textsc{Idnani, A.} (1983). A numerically
stable dual method for solving strictly convex quadratic programs.
\textit{Math. Program.} \textbf{27} 1--33.
\MR{0712108}

\bibitem[\protect\citeauthoryear{Hastie, Friedman and Tibshirani}{2001}]{hastie2001esl}
\textsc{Hastie, T., Friedman,~J.} and \textsc{Tibshirani, R.} (2001). \textit{The Elements
of Statistical Learning}. Springer, New York.
\MR{1851606}

\bibitem[\protect\citeauthoryear{Hastie et~al.}{2001}]{hastie2001supervised}
\textsc{Hastie, T., Tibshirani, R., Botstein, D.} and
\textsc{Brown,~P.} (2001). Supervised harvesting of expression trees.
\textit{Genome Biol.} \textbf{2} 0003--1.

\bibitem[\protect\citeauthoryear{Ishwaran et~al.}{2006}]{ishwaran2006random}
\textsc{Ishwaran, H., Kogalur, U., Blackstone, E.} and
\textsc{Lauer,~M.} (2006). Random survival forests. \textit{Ann. Appl.
Statist.} \textbf{2} 841--860.
\MR{2516796}

\bibitem[\protect\citeauthoryear{Johns et~al.}{2003}]{johns2003anthropogenic}
\textsc{Johns, T., Gregory, J., Ingram, W., Johnson, C., Jones, A.,
Lowe, J., Mitchell, J., Roberts, D., Sexton, D., Stevenson, D. et~al.}
(2003). Anthropogenic climate change for 1860 to 2100 simulated with
the HadCM3 model under updated emissions scenarios. \textit{Climate
Dynamics} \textbf{20} 583--612.

\bibitem[\protect\citeauthoryear{Liaw and Wiener}{2002}]{liawclassification}
\textsc{Liaw, A.} and \textsc{Wiener, M.} (2002). Classification and
regression by random forest. \textit{R News} \textbf{2} 18--22.

\bibitem[\protect\citeauthoryear{Lin and Jeon}{2006}]{lin2006random}
\textsc{Lin, Y.} and \textsc{Jeon, Y.} (2006). Random forests and
adaptive nearest neighbors. \textit{J. Amer. Statist. Assoc.}
\textbf{101} 578--590.
\MR{2256176}

\bibitem[\protect\citeauthoryear{Mangasarian, Street and
Wolberg}{1995}]{mangasarian1995breast}
\textsc{Mangasarian, O., Street, W.} and \textsc{Wolberg, W.} (1995).
Breast cancer diagnosis and prognosis via linear programming.
\textit{Oper. Res.} \textbf{43} 570--577.
\MR{1356410}

\bibitem[\protect\citeauthoryear{Meinshausen}{2009}]{meinshausen2009forest} \textsc{Meinshausen, N.} (2009). Forest Garrote.
\textit{Electron. J. Statist.} \textbf{3} 1288--1304.
\MR{2566188}

\bibitem[\protect\citeauthoryear{Nash et~al.}{1994}]{abalone}
\textsc{Nash, W., Sellers, T., Talbot, S., Cawthorn, A.} and
\textsc{Ford, W.} (1994). The population biology of abalone in
Tasmania. Technical report, Sea Fisheries Division.

\bibitem[\protect\citeauthoryear{Oakley and
O'Hagan}{2004}]{oakley2004probabilistic}
\textsc{Oakley, J.} and \textsc{O'Hagan, A.} (2004). Probabilistic
sensitivity analysis of complex models: A~Bayesian approach. \textit{J.
Roy. Statist. Soc. Ser. B} \textbf{66} 751--769.
\MR{2088780}

\bibitem[\protect\citeauthoryear{R Development Core Team}{2005}]{R}
\textsc{R Development Core Team} (2005). \textit{R: A Language and
Environment for Statistical Computing}. R Foundation for Statistical
Computing, Vienna, Austria.

\bibitem[\protect\citeauthoryear{Strobl et~al.}{2007}]{strobl2007bias}
\textsc{Strobl, C., ~Boulesteix, A., Zeileis, A.} and
\textsc{Hothorn,~T.} (2007). Bias in random forest variable importance
measures: Illustrations, sources and a solution. \textit{BMC
{B}ioinformatics} \textbf{8} 25.

\bibitem[\protect\citeauthoryear{Tibshirani}{1996}]{tibshirani96regression} \textsc{Tibshirani, R.} (1996). Regression shrinkage and
selection via the Lasso. \textit{J. Roy. Statist. Soc. Ser. B}
\textbf{58} 267--288.
\MR{1379242}

\bibitem[\protect\citeauthoryear{Wolpert}{1992}]{wolpert1992stacked}
\textsc{Wolpert, D.} (1992). Stacked generalization. \textit{Neural
{N}etworks} \textbf{5} 241--259.

\bibitem[\protect\citeauthoryear{Yu and B\"uhlmann}{2003}]{yu2003blr}
\textsc{Yu, B.} and \textsc{B\"uhlmann, P.} (2003). Boosting with the
L2 loss: Regression and classification. \textit{J.~Amer. Statist.
Assoc.} \textbf{98} 324--339.
\MR{1995709}

\end{thebibliography}
\end{document}